\documentclass[11pt,a4paper]{article}
\usepackage[hyperref]{acl2021}
\usepackage{times}
\usepackage{latexsym}

\usepackage{graphicx}
\usepackage{amsmath,amsfonts,amssymb}
\usepackage{mathtools}
\usepackage{caption}
\usepackage{lipsum}
\usepackage{booktabs}
\usepackage{color, colortbl}
\definecolor{Gray}{gray}{0.9}
\definecolor{ao(english)}{rgb}{0.0, 0.5, 0.0}
\definecolor{cardinal}{rgb}{0.77, 0.12, 0.23}
\usepackage{url}
\usepackage{float}
\usepackage{tabularx}
\usepackage{appendix}
\usepackage{placeins}
\usepackage{ctable}
\usepackage{multirow}
\usepackage{bm}
\usepackage{cancel}
\usepackage{hyperref}
\usepackage{capt-of}
\usepackage{cleveref}
\crefformat{section}{\S#2#1#3} 
\crefformat{subsection}{\S#2#1#3}
\crefformat{subsubsection}{\S#2#1#3}
\usepackage{microtype}
\usepackage[ruled,vlined]{algorithm2e}
\SetAlFnt{\small}
\SetAlCapFnt{\small}

\aclfinalcopy 

\title{Demoting the Lead Bias in News Summarization via Alternating Adversarial Learning}

\author{Linzi Xing\thanks{\hspace{0.05in} The first two authors contributed equally to this work.} , Wen Xiao\footnotemark[1] , Giuseppe Carenini\\
  Department of Computer Science \\
  University of British Columbia \\
  Vancouver, BC, Canada, V6T 1Z4 \\ 
  {\tt \{lzxing, xiaowen3, carenini\}@cs.ubc.ca}}

\date{}

\begin{document}
\maketitle
\begin{abstract}
In news articles the lead bias is a common phenomenon that usually dominates the learning signals for neural extractive summarizers, severely limiting their performance on data with different or even no bias.
In this paper, we introduce a novel technique\footnote{\url{https://github.com/lxing532/Debiasing}} 
to demote lead bias
and make the summarizer focus more on the content semantics.
Experiments on two news corpora with different degrees of lead bias show that our
method can effectively demote the model's learned lead bias and improve its generality on out-of-distribution data, with little to no performance loss on in-distribution data.
\end{abstract}

\section{Introduction}

Neural extractive summarization, which produces a short summary for a document by selecting a set of representative sentences, has shown great potential in real-world applications, including news \cite{cheng-lapata-2016-neural,summarunner} and scientific paper summarization \cite{cohan-etal-2018-discourse,xiao-carenini-2019-extractive}. 
Typically, a general-purpose extractive summarizer learns to select the most important sentences from a document to form the summary by considering their content salience, informativeness and redundancy. 
However, when restricted to a specific domain, the summarizer can learn to exploit particular biases in the data, the most famous of which is the \textit{lead bias} in news \cite{nenkova-etal-2011-automatic, hong-nenkova-2014-improving}; namely that sentences at the beginning of a news article are more likely to contain summary-worthy information.
As a result, not surprisingly, such bias is strongly captured by neural extractive summarizers for news, for which the sentence positional information tends to dominate the actual content of the sentence in model prediction \cite{jung-etal-2019-earlier, grenander-etal-2019-countering, zhong-etal-2019-searching, zhong-etal-2019-closer}.

While learning a summarizer reflecting the biases in the training dataset is completely fine when the summarizer is going to be deployed to summarize documents having similar biases, it would be problematic when the model was applied to deal with documents coming from a mixture of datasets with different degrees of such biases. 
In this paper, we address this problem in the context of the lead bias in the news domain by exploring ways in which an extractive summarizer for news can be trained so that it learns to balance the lead bias with the content of the sentences, resulting in a model that can be applied more effectively when the target documents belong to news datasets in which the lead bias is present in rather different degrees.

Recently, \citet{grenander-etal-2019-countering} proposes two preliminary solutions. One is to pretrain the summarizer on an automatic generated ``unbiased" corpus where the document sentences are randomly shuffled, which however has the negative effects of preventing the learning of inter-sentential information. The other, which can be only applied to RL-based summarizers, is to add an explicit auxiliary loss to directly balance position with content.  
Alternatively, \citet{zhong-etal-2019-closer} and \citet{Wang19} investigate strategies to train the summarizer on multiple news datasets with different degrees of lead bias, but this may still be problematic when we apply the trained summarizer to the documents with lead bias not covered in the training data.
Outside the summarization area, methods have also been proposed to eliminate data biases for other NLP tasks like text classification or entailment \cite{kumar-etal-2019-topics, clark-etal-2019-dont, clark-etal-2020-learning}.

Inspired by \citet{kumar-etal-2019-topics}, we have developed an alternating adversarial learning technique to demote the summarizer lead bias, but also maintain the performance on the in-distribution data. We introduce a position prediction component as an adversary, and optimize it along with the neural extractive summarizer in an alternating manner. 
Furthermore, in contrast with \citet{grenander-etal-2019-countering} and \citet{Wang19}, our proposal is model-independent and only requires one type of news dataset as training input.

In this paper, we apply our proposed method to a biased transformer-based extractive summarizer \cite{NIPS2017_7181} trained on CNN/DM training set \cite{NIPS2015_afdec700} and conduct experiments on two test sets with different degrees of lead bias: CNN/DM and XSum \cite{narayan-etal-2018-dont}, for in-distribution and generality evaluation respectively. The experimental results indicate that our proposed ``debiasing'' method can effectively demote the lead bias learned by the neural news summarizer and improve its generalizability, while still mostly maintaining the model’s performance on the data with a similar lead bias.

\section{Proposed Debiasing Method}
Our 
method aims 
to demote the lead bias learned by the summarizer and encourage it to select content based more on the semantics covered in sentences. As shown in Figure~\ref{fig:model}, our method comprises two components: one for \textit{Summarization} (red) and the other for sentence \textit{Position Prediction} (green).

\subsection{Summarization Component} \label{sec:summ_component}
\begin{figure}
\centering
\includegraphics[width=2.98in]{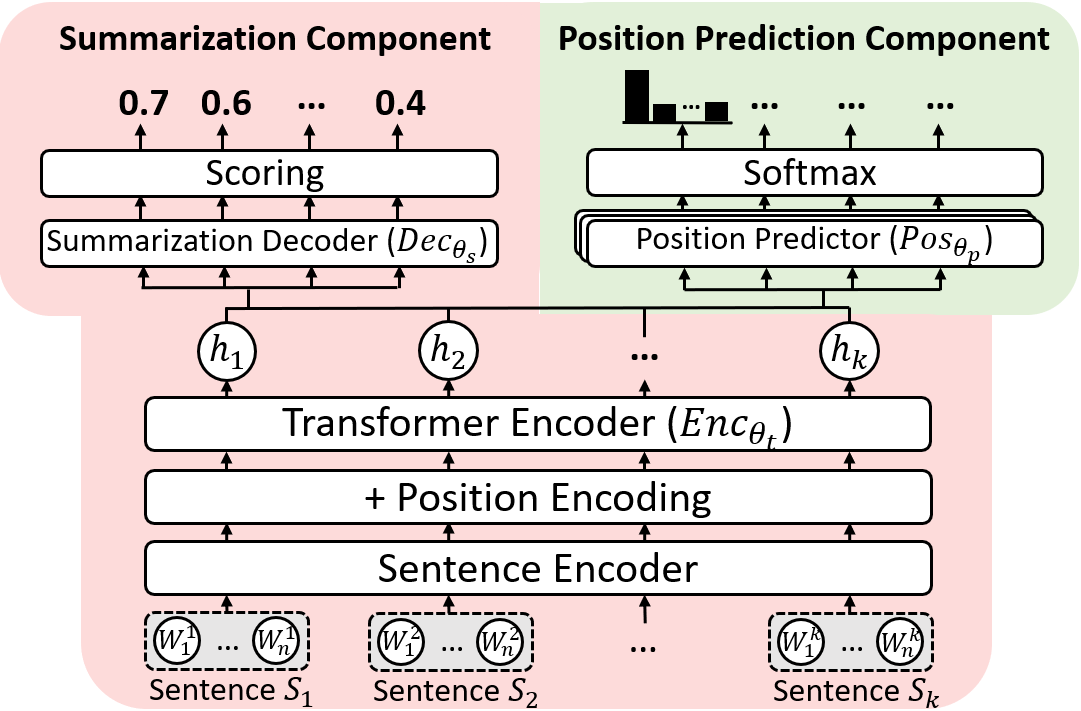}
\caption{\label{fig:model}
The overall architecture of our proposed lead bias demoting method. }
\vspace{-0.2in}
\end{figure}

Following previous work, we formulate extractive summarization as a sequence labeling task \cite{xiao-carenini-2019-extractive, xiao-carenini-2020-systematically, xiao-etal-2020-really}. For a document $d = \{ s_1, s_2, ... , s_k \}$, each sentence will be assigned a score $\alpha \in [0,1]$. The summary will then be formed with the highest scored sentences. We adopt a transformer-based model \cite{NIPS2017_7181} as our basic ``biased" summarization component (red in Fig.~\ref{fig:model}), as shown to be heavily impacted by the lead bias \cite{zhong-etal-2019-searching}.
This component contains a transformer-based encoder $Enc_{\theta_{t}}$ and a multilayer perceptron (MLP) decoder $Dec_{\theta_{s}}$, parameterized by $\theta_{t}$ and $\theta_{s}$ respectively. We use the averaged word embedding from Glove as sentence embedding as suggested in \citet{kedzie-etal-2018-content}. We optimize this summarization system by minimizing the loss:
\vspace{-0.1in}
\begin{equation}
\begin{aligned}
    L_1 &= -\frac{1}{N}\sum_{i=1}^{N} CE(\alpha_i, y_i) \label{eq:1} \\
    \alpha_i &= Dec_{\theta_{s}}(Enc_{\theta_t}(s_i))
\end{aligned}
\end{equation}
where $CE$ denotes the cross-entropy loss and $y_i \in \{0,1\}$ is the ground truth label for sentence $s_i$, representing if $s_i$ is selected to form the summary.

\subsection{Position Prediction Component} \label{sec:pos_component}
Our goal is to train the summarization model to make accurate predictions based more on the sentence semantics, rather than whether the sentence is in the lead position. More specifically, we aim to design an encoder network $Enc_{\theta_{t}}$ to output the set of contextualized sentence representations $\textbf{H} = \{h_1, ..., h_k\}$ which cover less sentence positional information, so that the following decoder $Dec_{\theta_{s}}$ will make predictions depending less on such positional information. To achieve this, the first step is to understand how much and in what form the positional information is encoded in $Enc_{\theta_{t}}$. 
Therefore, we propose a position prediction network to learn to predict the position of sentences in a document based only on $\textbf{H}$.
Intuitively, the higher accuracy this component can achieve, the more positional information is contained in $\textbf{H}$. This position prediction component will then play the role of an adversary module to demote the influence of lead bias presented in the training phase of the summarization component.

Concretely, because predicting the exact position for each sentence would require an extremely large set of labels with a skewed distribution, we choose to predict the portion of the document each sentence belongs to. In particular, once we obtain the set of contextualized sentence representations $\textbf{H}$ from the encoder network $Enc_{\theta_{t}}$, we initialize a MLP (parameterized by $\theta_{p}$ and followed by Softmax) as the position prediction component $Pos_{\theta_{p}}$ (green in Fig~\ref{fig:model}). In essence, this component $Pos_{\theta_{p}}$ takes $\textbf{H}$ as input and outputs a M-dimensional multinomial distribution for each sentence to represent its position in a document. More formally, $ Pos_{\theta_{p}}(h_i) = (\hat{p}^{(i)}_1, .., \hat{p}^{(i)}_j, .., \hat{p}^{(i)}_M)$ where $\sum_{j=1}^{M}\hat{p}^{(i)}_{j}=1$.
$\hat{p}^{(i)}_j$ is the predicted probability of the $i$th sentence belongs to the $j$th portion of a document when the document is divided into $M$ parts ($M$ is a tunable hyperparameter).  
We use the cross-entropy loss to optimize $Pos_{\theta_{p}}$ to extract sentence positional signals encoded in the system:
\begin{equation}
    L_2 = -\frac{1}{N}\sum_{i=1}^{N} CE(Pos_{\theta_{p}}(h_i), p_{i}) \label{eq:2}
\end{equation}
where $p_{i}$ is the true position of sentence $i$.

\subsection{Alternating Adversarial Learning}

To demote the influence of positional bias and balance it with the sentence semantics in the summarization system, we want to modify the encoder to produce $\textbf{H}$, which can still be accurate for summary generation but fail at sentence position prediction. We achieve this by alternatingly executing “Position learning” and “Position debiasing”, as proposed in \newcite{kumar-etal-2019-topics} and presented in Algorithm 1. In the “Position learning” phase, once a pretrained summarization system is obtained, we first fix its weights and train an adversary network $Pos^{*}_{\theta_{p}}$ (sentence position predictor) to extract the positional information contained in the encoder. Then in the “Position debiasing” phase, we fix the weights of $Pos^{*}_{\theta_{p}}$ and update the parameters of the summarization component to maximize the position prediction loss of adversary ($L_{adv}$ in eq~\ref{eq:3}) while minimizing the summarization loss $L_1$:
\vspace{-0.05in}
\begin{equation}
\begin{aligned}
    L_{3} &= \beta L_1 + (1-\beta) L_{adv} \label{eq:3} \\
    L_{adv} &= -\frac{1}{N}\sum_{i=1}^{N} CE(Pos^{*}_{\theta_{p}}(h_i), U_{M})
\end{aligned}
\end{equation}
To maximize the position prediction loss, the fixed adversary $Pos^{*}_{\theta_{p}}$ should ideally output the uniform distribution, $U_M = (\frac{1}{M}, ..., \frac{1}{M})$, for the position prediction of each sentence. $\beta$ is the trade-off parameter tuned at validation stage to control the degree of lead bias demoting.

In practice, we notice that reusing the same adversary for all iterations will make the positional signals not weakened but instead encoded in a different way. To avoid this problem, we follow \citet{kumar-etal-2019-topics} to use multiple adversaries (parameterized with $[\theta_{p}^{(1)}, ... ,\theta_{p}^{(N)}]$ in Algorithm 1), making it more difficult for the encoder to keep the sentence positional signals by encoding them into a more implicit format for position predictor to learn.
\begin{figure}
\centering
\includegraphics[width=3.02in]{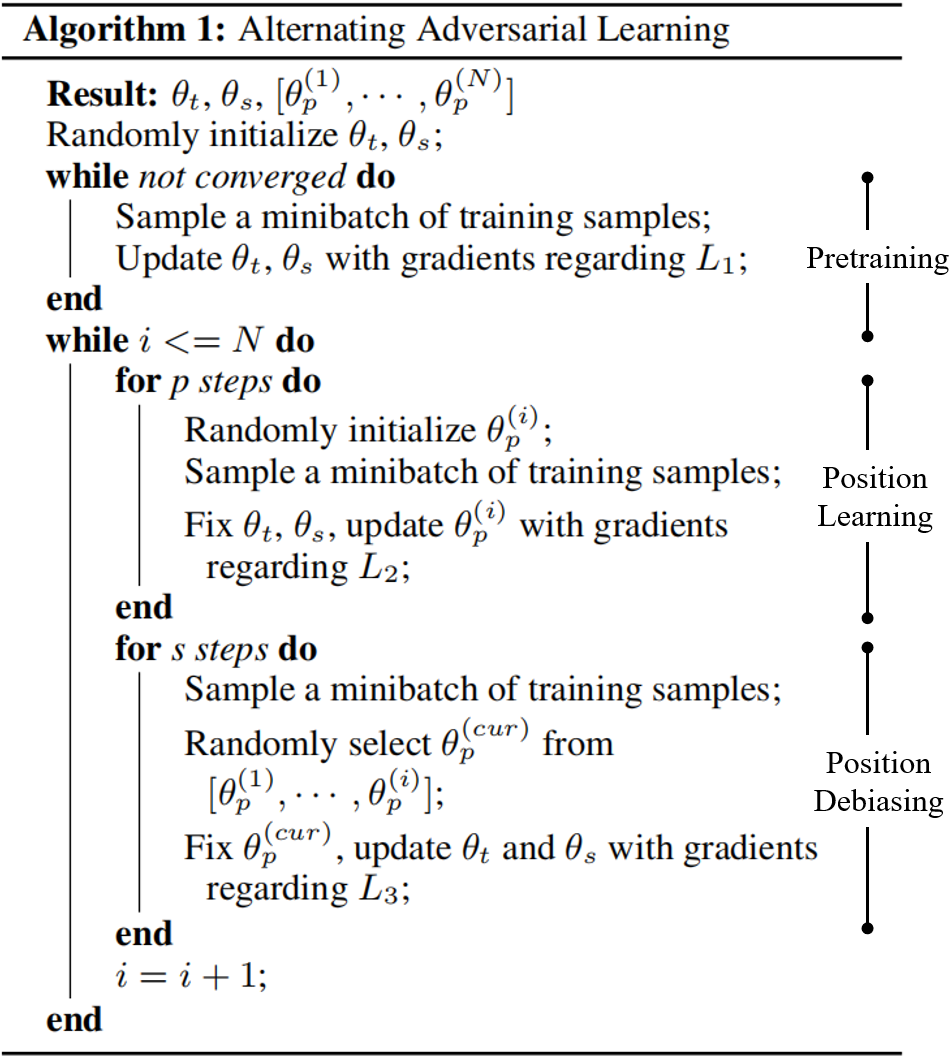}
\end{figure}

\begin{table*}
\centering
\scalebox{0.837}{
\begin{tabular}{c | l  l  l  l | l  l  l  l }
\specialrule{.1em}{.05em}{.05em}
\rowcolor{Gray}
\multicolumn{1}{c}{\textbf{Model}} & \multicolumn{4}{c}{\textbf{CNN/DM}} & \multicolumn{4}{c}{\textbf{XSum}}\\
\hline
  & \hspace{0.05in} R1 & \hspace{0.05in} R2 & \hspace{0.05in} RL & Mean & \hspace{0.05in} R1 & \hspace{0.02in} R2 & \hspace{0.04in} RL & Mean \\
\hline
Lead & 40.30 & 17.52 & 36.54 & 31.45 & 16.32 & 1.60 & 11.96 & 9.96 \\
Oracle & 56.04 & 33.10 & 52.29 & 47.14 & 30.98 & 8.98 & 23.51 & 21.16 \\
\hline
Basic Transformer \cite{NIPS2017_7181} & 41.02 & 18.39 & 37.39 & 32.27 & 16.79 & 1.84 & 12.33 & 10.32 \\
\textit{\small -- No Position Encoding} & 37.82$\downarrow$ & 15.59$\downarrow$ & 34.32$\downarrow$ & 29.24 & \textbf{18.29}$\uparrow$ & \textbf{2.53}$\uparrow$ & \textbf{13.45}$\uparrow$ & \textbf{11.42}  \\
\hspace{0.25cm}\textit{\small -- Only Position Encoding} & 40.13$\downarrow$ & 17.36$\downarrow$ & 36.38$\downarrow$ & 31.29 & 16.22$\downarrow$ & 1.62$\downarrow$ & 11.90$\downarrow$ & 9.91  \\
\hline
Learned-Mixin \cite{clark-etal-2019-dont} & 40.72$\downarrow$ & 18.27 & 37.17$\downarrow$ & 32.05 & 16.67 & 1.91$\uparrow$ & 12.28 & 10.29 \\
Shuffling \cite{grenander-etal-2019-countering} & \textbf{41.00} & \textbf{18.43} & \textbf{37.37} & \textbf{32.27} & 16.98$\uparrow$ & 1.96$\uparrow$ & 12.48$\uparrow$ & 10.47 \\
Our Method & \underline{40.88}$\downarrow\Downarrow$ & \underline{18.37} & \underline{37.27}$\downarrow$ & \underline{32.18} & \underline{17.20}$\uparrow\Uparrow$ & \underline{1.99}$\uparrow\Uparrow$ & \underline{12.63}$\uparrow\Uparrow$ & \underline{10.61} \\
\specialrule{.1em}{.05em}{.05em}
\end{tabular}}
\caption{\label{tab:res_cnn_xsum} The ROUGE-1/2/L F1 scores and ``Mean" (mean of ROUGE-1/2/L) on CNN/DM and XSum test data. The best and second best performances over the basic transformer are in \textbf{bold} and \underline{underlined}. $\uparrow$/$\downarrow$ indicates the results are significantly higher/lower than Basic Transformer and $\Uparrow$/$\Downarrow$ indicates the results are significantly higher/lower than Shuffling ($p<0.01$ with bootstrap resampling test \cite{lin-2004-rouge}).}
\end{table*}

\begin{table}
\centering
\scalebox{0.85}{
\begin{tabular}{c | l  l  l }
\specialrule{.1em}{.05em}{.05em}
\rowcolor{Gray}
\multicolumn{1}{c}{\textbf{Model}} & \multicolumn{1}{l}{\textbf{D$_{early}$}} & \multicolumn{1}{l}{\textbf{D$_{middle}$}} & \multicolumn{1}{l}{\textbf{D$_{late}$}} \\
\hline
Lead-3 & 49.33 & 30.90 & 19.80 \\
Oracle & 49.51 & 47.02 & 43.81 \\
\hline
Basic Transformer & 44.30 & 31.91 & 22.65 \\
\textit{\small -- No Position Encoding} & 16.07 & 16.88 & 18.59 \\
\textit{\small -- Only Position Encoding} & 48.65$^{*\dagger\ddagger}$ & 30.97 & 19.70 \\
\hline
Learned-Mixin & 40.45 & 31.82 & 22.70 \\
Shuffling & 42.69 & 31.91 & 22.99$^{*\dagger\ddagger}$ \\
Our Method & 42.67 & 32.18$^{*\dagger\ddagger}$ & 22.85$^{*\dagger}$ \\
\specialrule{.1em}{.05em}{.05em}
\end{tabular}
}
\caption{\label{tab:res_eml} Avg. of ROUGE-1/2/L F1 scores on $D_{early}$, $D_{middle}$ and $D_{late}$. Results significantly better than Basic Transformer on ROUGE-1/2/L are marked with $*$, $\dagger$, and $\ddagger$ respectively.
}
\end{table}

\vspace{-0.15in}
\section{Experiments and Analysis}

\subsection{Datasets}
\label{sec:datasets}
We use the standard CNN/DM dataset (204,045 training, 11,332 validation and 11,334 test data) \cite{NIPS2015_afdec700} for training since it is one of the mainstream news datasets with observed lead bias \cite{jung-etal-2019-earlier, grenander-etal-2019-countering}. For model evaluation, we use the test set of CNN/DM to evaluate model's in-distribution performance, as well as the test set of XSum \cite{narayan-etal-2018-dont}, which consists of 11,334 datapoints, to evaluate model's generality when transferred to less biased data. The empirical analysis in \citet{narayan-etal-2018-dont} and \citet{jung-etal-2019-earlier} shows the documents and summaries in XSum are shorter and have less lead bias compared to CNN/DM.

\subsection{Experimental Design}
{\bf Baselines:}
We compare our proposal with various baselines (see Table~\ref{tab:res_cnn_xsum}). The top section of Table~\ref{tab:res_cnn_xsum} presents \textit{\textbf{Lead}} baseline and \textit{\textbf{Oracle}}. For CNN/DM, lead baseline refers to \textit{\textbf{Lead-3}} and for XSum, it refers to \textit{\textbf{Lead-1}}. The middle section of Table~\ref{tab:res_cnn_xsum} contains the basic transfomer-based summarizer accepting ``sentence representation + position encoding" as input, and its two variants, one without positional encoding, while the other with only positional encoding as input. The bottom section contains \textit{\textbf{Shuffling}} \cite{grenander-etal-2019-countering}, which is a method proposed lately for summarization lead bias demoting, and \textit{\textbf{Learned-Mixin}} \cite{clark-etal-2019-dont}, which is a general debiasing method proposed to deal with NLP tasks when the type of data bias in the training set is known and bias-only model is available. In our case, the data bias is lead bias and the bias-only model is the transformer trained with only positional encoding as input.

\vspace{0.4ex}
\noindent
{\bf Implementation Details:} All the transformer-based models have the same setting as the standard transformer \cite{NIPS2017_7181}, with 6 layers, 8 heads per layer, and $d_{model}=512$. We use Adam to train all the models with scheduled learning rate with warm-up (initial learning rate $lr=2e-3$). We choose the top-3 sentences to form the final summary for CNN/DM and the top-1 sentence for XSum due to the different average summary lengths. 
The class number of sentence position $M$ is set to 10 and trade-off parameter $\beta$ is set to 0.9 (searched from 0 to 1, by increasing 0.1 for each step). 
We tune these hyper-parameters on a ``balanced" validation set sampled from the standard CNN/DM validation data.

\begin{figure*}
\centering
\begin{minipage}[t]{0.5\textwidth}
\centering
\includegraphics[width=0.98\textwidth]{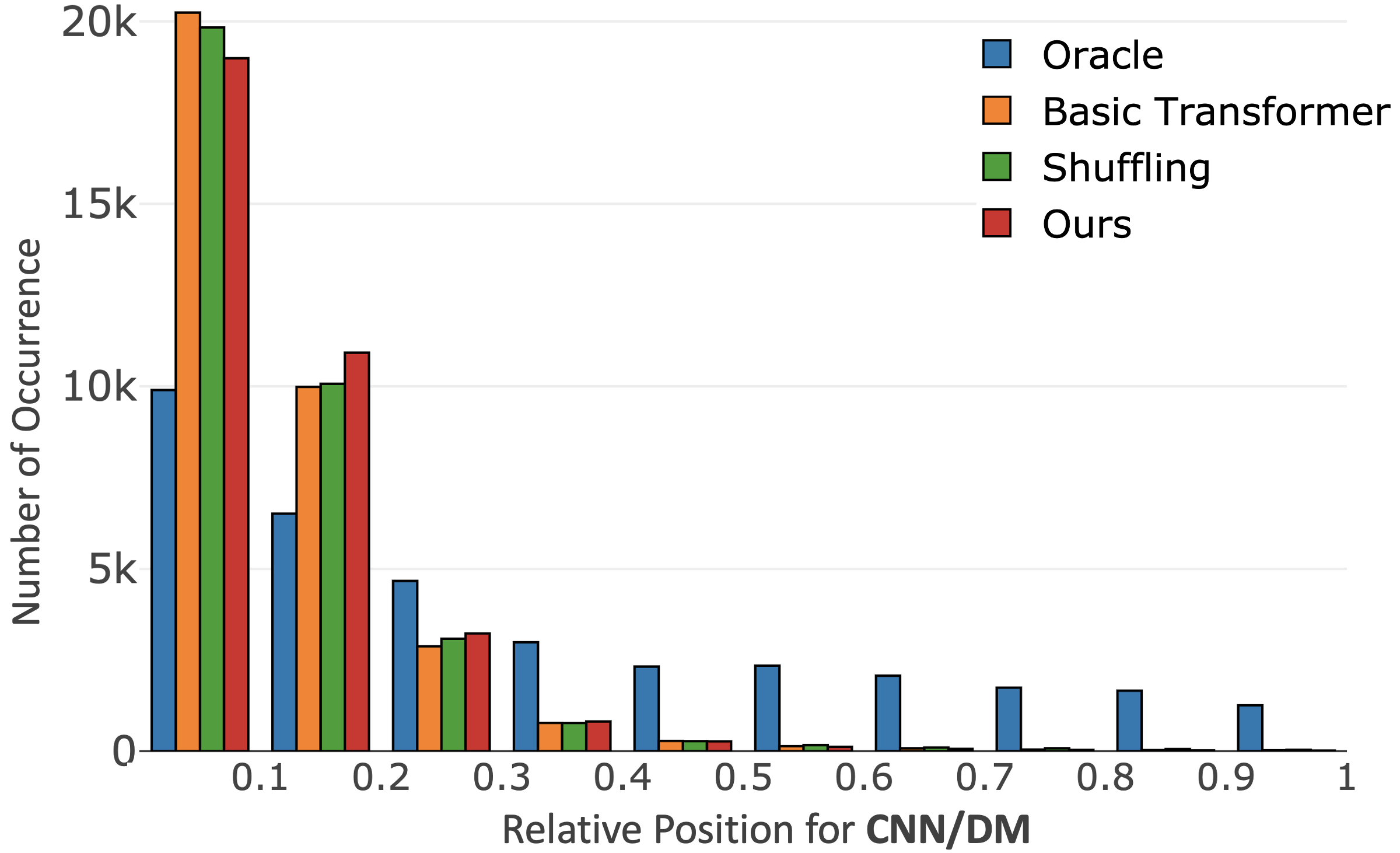}
\label{fig:side:a}
\end{minipage}%
\begin{minipage}[t]{0.5\textwidth}
\centering
\includegraphics[width=0.98\textwidth]{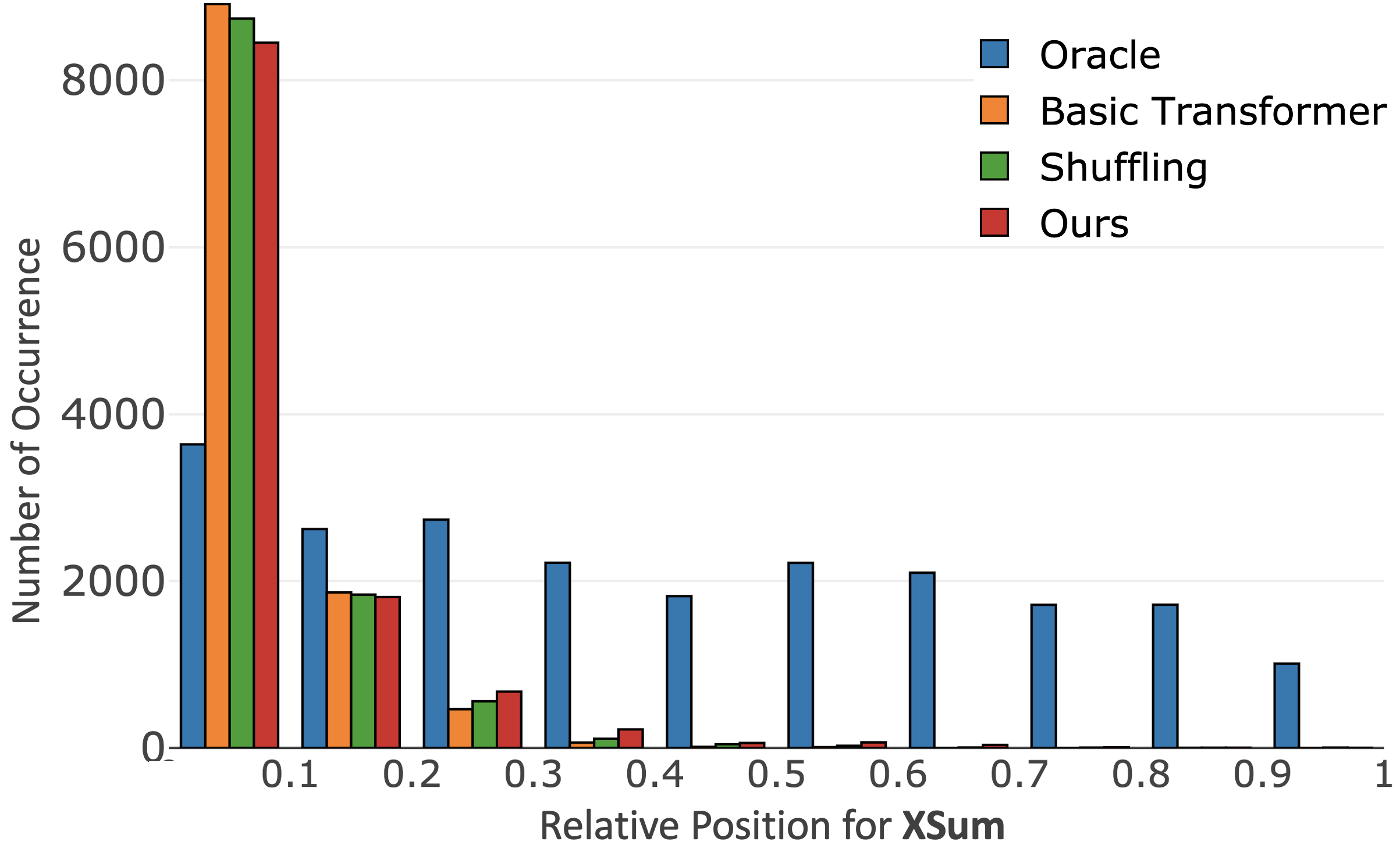}
\label{fig:side:b}
\end{minipage}
\caption{\label{fig:pos_distribution} 
Relative position distributions of selected sentences in the original document of two testing corpora (CNN/DM and XSum), obtained by different lead bias demoting strategies.}
\end{figure*}

\subsection{Results and Analysis}
Table~\ref{tab:res_cnn_xsum} reports the performance of the chosen baselines and our proposal on CNN/DM test set, which has the same data distribution as the training data, and XSum test set, which is from another news resource and with much less lead bias than CNN/DM.

From the middle section of Table~\ref{tab:res_cnn_xsum}, we observe that if we withhold the position cues (\textit{No Position Encoding}) by using only semantic representation as input, the model's performance drops considerably on CNN/DM, but remarkably increase on XSum. In contrast, if we merely use position cues as input (\textit{Only Position Encodings}), the decrease of the performance on CNN/DM becomes much more modest, while there is  substantial performance drop on XSum. These results confirm that positional signal is a rather important feature for bias-relied neural summarizers. However, relying too much on it will also limit model's generality when applied to the dataset with less bias than the training samples. 
Therefore, seeking strategies to balance the semantics and position features is crucial for the neural extractive summarization for news.

When we compare the lead bias demoting methods presented at the bottom 
of Table~\ref{tab:res_cnn_xsum}, our proposal and  \textit{Shuffling} give significant performance boosting on XSum, while \textit{Learned-Mixin} results in performance decrease on both datasets. Comparing our method and \textit{Shuffling} directly, while they are essentially equivalent on maintaining the performance on the in-distribution CNN/DM data (0.09 difference in terms of the average of ROUGE scores (ROUGE-Mean)), our method provides a significant improvement on XSum, and outperforms \textit{Shuffling} and the basic transformer by 0.14 and 0.29 on ROUGE-Mean respectively. It is noteworthy that the transformer without position encoding achieves the best performance on XSum. However, it is the worst system on in-distribution data. Throughout all the comparisons, our proposal can best balance the sentence position and content semantics. 

To more deeply investigate the behavior of our demoting method on the documents whose summary sentences are from different document portions, we follow \citet{grenander-etal-2019-countering} to create three subsets, $D_{early}$, $D_{middle}$, $D_{late}$, from the CNN/DM testing set. Documents are ranked by the mean of their summary sentences' indices in ascending order, and then the top-ranked 100 documents, the 100 documents closest to the median, and the bottom-ranked 100 documents are selected to form $D_{early}$, $D_{middle}$, $D_{late}$\footnote{
Due to the common generation mechanism of oracles, the number of sentences in the oracle is not fixed. For fair comparison, we only consider adding documents with the oracle having exactly 3 sentences into $D_{early}$, $D_{middle}$, $D_{late}$.}.
Results in Table~\ref{tab:res_eml} show that even if our model does not match the basic transformer on documents in $D_{early}$, it does yield benefits for both $D_{middle}$ and $D_{late}$ with significant improvements, while the competitive baseline \textit{Shuffling} only achieves that on $D_{late}$.

\vspace{0.5ex}
\noindent
{\bf Position of Selected Content:} To more explicitly investigate how well the prediction of different models fits the ground-truth sentence selection (\textit{Oracle}), we compare the relative position of the selected content of our method with the undebiased model (\textit{Basic Transformer}) and the most competitive debiased model (with \textit{Shuffling}), as illustrated in Figure~\ref{fig:pos_distribution}. We can observe that: (1) CNN/DM contains much more lead bias than XSum, shown by a more right-skewed histogram for \textit{Oracle}. Thus, the basic transformer trained on it is also heavily impacted by the lead bias and tends to select sentences $\in [0, 0.1]$ with much higher probability even on the less biased XSum. (2) While \textit{Shuffling} and our method can both effectively demote the extreme trend towards selecting sentences in the lead position, our method seems to be sightly better at encouraging the model to select sentences with higher relative position.


\section{Conclusion and Future Work}
We propose a lead bias demoting method to make news extractive summarizers more robust across datastets, by optimizing a position prediction and a summarization component in an alternating way.
Experiments indicate that our 
method 
improves model's generality on out-of-distribution data, while still largely maintaining its performance on in-distribution data.
As such, it represents the best viable solution when at inference time input documents may come from an unknown mixture of datasets with different degrees of position bias.

For the future, we plan to explore more sophisticated and effective methods (e.g., adjusting the lead bias online) and infuse them together with neural abstractive summarization models, known to generate more succinct and natural summaries. Another interesting direction for future work can be 
exploring the potential of applying our proposed bias demoting strategy to other tasks, which can also be framed as the sequence labeling problem and possibly troubled by biases in the training data (e.g., Topic Segmentation \cite{xing-etal-2020-improving} and Semantic Role Labeling \cite{ouchi-etal-2018-span}). 

\section*{Acknowledgments}
\vspace{-1mm}
We thank the anonymous reviewers and the UBC-NLP group for their insightful comments and suggestions. This research was supported by the Language \& Speech Innovation Lab of Cloud BU, Huawei Technologies Co., Ltd.

\bibliographystyle{acl_natbib}
\bibliography{anthology,acl2021}


\end{document}